\documentclass{article}
\usepackage{iclr2021_conference,times}
\iclrfinalcopy 

\usepackage[utf8]{inputenc}
\usepackage[T1]{fontenc}

\usepackage[hyphens]{url}
\usepackage{hyperref}
\usepackage{graphicx}
\hypersetup{
    colorlinks,
    citecolor=blue,
    filecolor=blue,
    linkcolor=blue,
    urlcolor=blue,
}

\usepackage{natbib}
\setcitestyle{authoryear,open={(},close={)}}

\usepackage{amsmath,amssymb}
\usepackage{makecell}
\usepackage{chngcntr} 

\usepackage{caption}
\usepackage{xcolor}

\newcommand{\future}{(T+1):(T+F)}
\newcommand{\past}{1:T}
\newcommand{\ypast}{y_{\past}}
\newcommand{\yfuture}{y_{\future}}
\newcommand{\fpast}{f_{\past}}
\newcommand{\ffuture}{f_{\future}}
\newcommand{\fh}{f^{h}}
\newcommand{\yh}{y^{h}}
\newcommand{\fsh}{f^{s,h}}
\newcommand{\ysh}{y^{s,h}}
\newcommand{\lambdah}{\lambda^h}
\newcommand{\lambdash}{\lambda^{s,h}}

\title{Forecasting COVID-19 Counts At A Single Hospital: A Hierarchical Bayesian Approach}
\date{February 2021}
\author{Alexandra Hope Lee \\
    Tufts University \\
\And
    Panagiotis Lymperopoulos \\
    Tufts University \\
\And
    Joshua T. Cohen \\
    Tufts Medical Center \\
\And
    John B. Wong \\
    Tufts Medical Center \\
\And
    Michael C. Hughes \\
    Dept. of Computer Science, Tufts University \\
}

\begin{document}

\setlength{\abovedisplayskip}{2pt plus 3pt}
\setlength{\belowdisplayskip}{2pt plus 3pt}

\maketitle

\begin{abstract}
We consider the problem of forecasting the daily number of hospitalized COVID-19 patients at a single hospital site, in order to help administrators with logistics and planning. We develop several candidate hierarchical Bayesian models which directly capture the count nature of data via a generalized Poisson likelihood, model time-series dependencies via autoregressive and Gaussian process latent processes, and can share statistical strength across related sites.
We demonstrate our approach on public datasets for 8 hospitals in Massachusetts, U.S.A. and 10 hospitals in the United Kingdom.
Further prospective evaluation compares our approach favorably to baselines currently used by stakeholders at 3 related hospitals to forecast 2-week-ahead demand by rescaling state-level forecasts.

\end{abstract}

\section{Introduction}
The COVID-19 pandemic has created unprecedented demand for limited hospital resources across the globe. Emergency resource allocation decisions made by hospital administrators (such as planning additional personnel or provisioning beds and equipment) are crucial for achieving successful patient outcomes and avoiding overwhelmed capacity. However, at present hospitals often lack the ability to forecast what will be needed at their site in coming weeks. This may be especially true in under-resourced hospitals, due to constraints on funding, staff time and expertise, and other issues.
In response to this pressing need, in this study our goal is to develop statistical machine learning approaches to forecast hospital utilization for specific hospitals. By focusing on the short-term future (1-3 weeks ahead) at specific sites, we hope predictions are directly \emph{actionable} so hospital administrators can respond to forecasted demand.

While many efforts to forecast the spread of COVID-19 and its impact on hospitals have been publicized~\citep{murrayForecastingImpactFirst2020,jewellPredictiveMathematicalModels2020,reinerModelingCOVID19Scenarios2021},
they are not usable for off-the-shelf predictions for a specific hospital because they focus on whole countries, states, or regions rather than a specific site.
Even if this regional forecasting were reliable, it can be of limited relevance to a particular hospital~\citep{wongPandemicSurgeModels2020}, at which new patient arrivals depend on localized conditions (e.g., case incidence at local ``hot spots'').
Notable efforts for hospital-level or patient-level modeling exist~\citep{epsteinPredictiveModelPatient2020,roimiDevelopmentValidationMachine2021}, but require much more detailed data than our approach (e.g. length-of-stay for all patients at a site or other patient-level covariates).

\textbf{Contributions.} Our study develops and validates latent variable models to predict \emph{distributions} over census counts at a hospital site for each future day of interest given a univariate time series of past counts. Our probabilistic methods can work even if the past census data is not fully observed (as might arise in sites relying on noisy or error-prone record-keeping processes). As a technical contribution, we show that generalized Poisson likelihoods are a better alternative to the more popular standard Poisson likelihoods for modeling hospitalization census counts. We further show that when modeling multiple hospital sites in the same region, we can use hierarchical modeling to \emph{share} statistical strength and improve forecasts. We have released open-source Python code\footnote{\url{https://github.com/tufts-ml/single-hospital-count-forecasting/}} to allow others to reproduce our analysis, making site-specific demand prediction at sites around the globe possible.


\section{Methods}
Suppose across $T$ days we observe a univariate time series $y_{1:T} = [y_1, y_2, \ldots y_T]$, where $y_t \in \{0, 1, 2, \ldots \}$ indicates the patient occupancy count for a specific hospital on day $t$.
Our goal is to develop \emph{forecasts} for the next $F$ days (typically a few weeks ahead), given all previous observations, using a conditional probabilistic model $p( y_{(T+1):(T+F)} \mid y_{1:T} )$.
We consider a flexible family of latent variable models, where each timestep has a latent real value $f_t \in \mathbb{R}$. We capture dependency across time in the latent series $f_{1:T}$, and model each count $y_t$ as conditionally independent given $f_t$:
\begin{align}
    p(f_{1:T}, y_{1:T} ) = p_{\alpha}(f_{1:T}) \cdot \textstyle \prod_{t=1}^T p_{\lambda}(y_t \mid f_{t} ), \label{eq:latent-variable-time-series}
\end{align}
with latent-generating parameters $\alpha$ and count-generating parameters $\lambda$.
We'll consider two options for $p_{\alpha}$ motivated by different dependency assumptions, 
as well as two possible likelihoods $p_{\lambda}$.

\textbf{Likelihoods for count data.}
Given $f_t$, to generate the observed count $y_t$ on day $t$, we use either the standard Poisson or the generalized Poisson distribution.
In both cases, we set the mean parameter by transforming the latent $f_t$ to a positive value via the exponential:
\begin{align}
	 p_{\lambda}(y_t \mid f_t) &= \text{Poisson}( \exp(f_t) ),
	 ~~\text{or}~~ p_{\lambda}(y_t \mid f_t) = \text{GenPoisson}( \exp(f_t), \lambda )
\end{align}
The standard Poisson has no ability to control variance separately from the mean.
The generalized Poisson~\citep{consulGeneralizedPoissonRegression1992} has dispersion parameter: $\lambda \in [-1, +1]$, reducing to the standard Poisson when $\lambda = 0$.
We assume a priori symmetric chances of both under- and over-dispersion, so we set the prior on $\lambda$ be a normal distribution with mean 0 and standard deviation 0.3, truncated to $[-1, +1]$. Typically, on our hospital data we find that our posteriors favor under-dispersion.

\textbf{Generalized Autoregressive (GAR) model.}
Our generalized autoregressive model is an instance of the generative model in Eq.~\eqref{eq:latent-variable-time-series} with an order-$W$ autoregressive process to generate $f_{1:T}$:
\begin{align}
    p_{\alpha}(f_{1:T}) = \textstyle \prod_{t=1}^T
    	p_{\alpha}( f_t \mid f_{(t-W):(t-1)} )
    = \textstyle \prod_{t=1}^T \mathcal{N}( f_t | \beta_0 + \sum_{\tau = 1}^{W} \beta_{\tau} f_{t-\tau}, \sigma^2)
\end{align}
We place vague unimodal priors over the parameters $\alpha = \{ \beta, \sigma \}$ (see App.~\ref{sec:methods-single-site-GAR}).
Window size $W$ is a hyperparameter selected on validation data.
Our GAR model is limited to linear dependencies within the latent sequence $f_{1:T}$, but it is conceptually simple and fast to fit and evaluate.

\textbf{Generalized Gaussian Process (GGP) model.}
We next consider a model where latents are drawn from a Gaussian process (GP), which can be seen as a simplified GP state-space model~\citep{damianouVariationalGaussianProcess2011,frigolaVariationalGaussianProcess2014}.
The goal here is a flexible, data-driven model for non-linear trends in $f_{1:T}$ without an explosion of parameters to learn.
Our model is again an instance of Eq.~\eqref{eq:latent-variable-time-series}, with a GP for the prior over the latents:
$f_{1:T} \sim \text{GP}( m_{\alpha}(t), k_{\alpha}(t,t') )$. We assume a constant mean $m_{\alpha}(t) = c$ and squared exponential covariance kernel $k_{\alpha}(t, t') = a^2 \exp\left(-\frac{(t - t')^2}{2 \ell^2}\right)$.
The parameters $\alpha = \{c, a, \ell\}$, with $a > 0$ and $\ell > 0$, are given vague unimodal priors (see App.~\ref{sec:methods-single-site-GGP}).

\textbf{Multi-site hierarchical model.}
Now, consider predicting future census counts at $H$ different hospital sites simultaneously, given the same $T$ days of observations from each site. If all $H$ sites share common trends (i.e. cases rising in all because they draw from similar populations), we might improve forecasts by modeling sites in the hierarchical Bayesian fashion~\citep{gelmanMultilevelHierarchicalModeling2006}.

Our multi-site generative model ties the latent-sequence-generating parameters $\alpha$ across sites, but allows likelihood parameters $\lambdah$ to be specific to each site (indexed by $h$). The model factorizes as:
\begin{align}
    p( \{ \yh_{1:T}, \fh_{1:T} \}_{h=1}^H )
    &= \textstyle \prod_{h=1}^H p_{\alpha}( \fh_{1:T} ) p_{\lambdah}( \yh_{1:T} \mid \fh_{1:T} ).
\end{align}
For simplicity, all multi-site experiments use the GAR model described above, with its latent-generating parameters $\alpha = \{ \beta, \sigma \}$ shared among all sites, again with vague priors (App.~\ref{sec:methods-multi-site}).

\textbf{Posterior estimation and forecasting.}
For all methods, we use a No-U-Turn sampler~\citep{hoffmanNoUTurnSamplerAdaptively2014} to perform Markov chain Monte Carlo approximate sampling from the posterior, as implemented using the PyMC3 toolbox~\citep{salvatierProbabilisticProgrammingPython2016}.
This lets us sample from the posterior over parameters and latent values: $p(\alpha, \lambda, f_{1:T} \mid y_{1:T})$ for single site models and $p(\alpha, \{\lambdah, \fh_{1:T} \}_{h=1}^H \mid \{\yh_{1:T} \}_{h=1}^H)$ for multi-site models, gathering $S$ posterior samples (indexed by $s$). We collect thousands of samples from multiple chains to avoid poor convergence.

Then, we can sample $S$ forecasts for site $h$ by conditioning on each posterior sample's parameters $\alpha^s, \lambdash$ and latents $\fsh_{1:T}$ when simulating the next $F$ days (indexed by $\tau$) from the model:
\begin{align}
    \fsh_{T+\tau} &\sim  p_{\alpha^s}( \fh_{T+\tau} \mid \fsh_{1:T+\tau-1} ), \quad
    \ysh_{T+\tau} \sim p_{\lambdash}( \yh_{T+\tau} \mid \fsh_{T+\tau} ), \quad \tau \in 1, 2, \ldots F.
\end{align}
After drawing $S$ forecast samples, we can compute summary statistics of these samples, such as mean or median values as well as lower and upper percentiles.

\section{Evaluation}
We applied our models to the task of forecasting site-specific census counts of patients with COVID-19.
First, we perform a retrospective  evaluation of our proposed models on two public datasets from April to July 2020 (described in App.~\ref{supp:datasets}).
These initial experiments consider data from 8 hospitals in Suffolk County, MA, U.S.A, as well as 10 hospital sites in the UK.
Next, we applied the best performing models in a \emph{prospective} validation on data from January to February 2021, comparing our methods to those currently used by stakeholders at a major hospital system in Massachusetts.

\textbf{Comparison of likelihoods: Generalized vs. Standard Poisson.}
Supplementary Fig.~\ref{fig:poisson_vs_genpoisson} shows that our proposed generalized Poisson likelihood delivers better heldout likelihoods on the MA data than the standard Poisson. We use the generalized Poisson throughout the remaining experiments.

\textbf{Comparison of models: Single-Site GGP vs Single-Site GAR vs Multi-Site GAR.}
Table~\ref{tab:mass_data_quantitative_results} provides quantitative heldout likelihood comparisons of our three candidate methods  (single-site GGP, single-site GAR, and multi-site GAR), across the 8 sites in MA and 10 sites in the UK.
We find that the multi-site GAR model is preferred for 6 out of 8 sites in the MA dataset, perhaps because all sites come from the same county and thus sharing information across sites works well.
The UK hospital sites are much more spread out geographically, so we see less conclusive results (multi-site GAR competes well in a plurality of 4 out of 10 sites).
Figure~\ref{fig:mass_data_qualitative_forecasts} illustrates forecasts for 3 representative sites from MA, showing that the GAR models extrapolate trends better than the GGP.

\begin{table}[!t]
\scriptsize
\begin{center}
\begin{tabular}{|p{3.5cm}| p{2.2cm}| p{2.2cm}| p{2.2cm}|} 
 \hline
 \textbf{Hospital Site (MA, USA)}
 & \textbf{Single-Site} \textbf{GGP}
 & \textbf{Single-Site} \textbf{GAR}
 & \textbf{Multi-Site} \textbf{GAR} \\
 \hline
 Beth Israel Deaconness
 	& $-3.153 \pm 0.005$ ~
 	& $-3.121 \pm 0.017$ ~
 	& $-3.010 \pm 0.011$ * \\
 \hline
 Boston Medical Center
 	& $-3.376 \pm 0.002$ ~
 	& $-3.454 \pm 0.025$ ~
 	& $-3.296 \pm 0.014$ * \\
 \hline
Brigham \& Women's Faulkner 
	& $-2.719 \pm 0.002$ ~
	& $-2.684 \pm 0.010$ ~
	& $-2.601 \pm 0.008$ * \\
 \hline
Brigham \& Women’s Hospital
	& $-3.118 \pm 0.001$ ~
	& $-2.978 \pm 0.009$*
	& $-3.044 \pm 0.016$ ~ \\
 \hline
 Carney Hospital
 	& $-3.072 \pm 0.008$  ~
 	& $-2.721 \pm 0.105$  ~
 	& $-2.564 \pm 0.090$  * \\
 \hline
Massachusetts General
	& $-3.777 \pm 0.020$ ~
	& $-3.468 \pm 0.018$  *
	& $-3.450 \pm 0.017$ * \\
 \hline
St. Elizabeth’s
	& $-1.980 \pm 0.007$ ~
	& $-1.312 \pm 0.120$ *
	& $-1.598 \pm 0.007$ ~ \\
 \hline
 Tufts Medical Center
 	& $-2.937 \pm 0.005$ ~
 	& $-2.822 \pm 0.010$ ~
 	& $-2.728 \pm 0.007$ * \\
 \hline
\end{tabular}
\begin{tabular}{|p{3.5cm}|p{2.2cm}|p{2.2cm}|p{2.2cm}|} 
\hline
 \textbf{Hospital Site (UK)}
 & \textbf{Single-Site} \textbf{GGP}
 & \textbf{Single-Site} \textbf{GAR}
 & \textbf{Multi-Site} \textbf{GAR} \\
 \hline
 Barts Health
 	&  $-4.730 \pm 0.058$  *
 	&  $-6.727 \pm 0.221$  ~
 	&  $-4.842 \pm 0.078$ ~ \\
 \hline
 Chelsea \& Westminster 
 	&  $-2.937 \pm 0.009$  *
 	&  $-3.141 \pm 0.030$ ~
 	&  $-5.929 \pm 0.086$  ~ \\
 \hline
 Imperial College Healthcare
 	&  $-4.017 \pm 0.019$  ~
 	&  $-3.063 \pm 0.178$  *
 	&  $-3.976 \pm 0.110$ ~ \\
 \hline
 King's College Hospital
 	&  $-3.012 \pm 0.003$ ~
 	&  $-2.749 \pm 0.020$ ~
 	&  $-2.695 \pm 0.005$  *\\
 \hline
 London North West University
 	&  $-2.829 \pm 0.008$ ~
 	&  $-2.488 \pm 0.014$  *
 	&  $-2.572 \pm 0.009$ ~ \\
 \hline
 Manchester University
 	&  $-12.335 \pm 0.647$  ~
 	&  $-12.735 \pm 0.937$  ~
 	&  $-6.152 \pm 0.167$ * \\
 \hline
 North Middlesex University
 	&  $-1.593 \pm 0.008$  *
 	&  $-1.677 \pm 0.022$  ~
 	&  $-1.703 \pm 0.008$ ~ \\
 \hline
 Nottingham University
 	&  $-3.443 \pm 0.018$  ~
 	&  $-3.324 \pm 0.066$  ~
 	&  $-2.950 \pm 0.032$ * \\
 \hline
 University Hospitals Birmingham
 	&  $-4.385 \pm 0.005$ *
 	&  $-4.571 \pm 0.019$ ~
 	&  $-4.583 \pm 0.031$  ~\\
 \hline
 Univ. Hospitals of North Midlands
 	&  $-3.188 \pm 0.002$ ~
 	&  $-2.953 \pm 0.033$ ~
 	&  $-2.829 \pm 0.016$ * \\
 \hline
\end{tabular} 
\end{center}

\caption{\textbf{Heldout likelihood for retrospective evaluation across 8 sites in MA (top) and 10 sites in UK (bottom).} We report test-set log likelihood (normalized by total test days). The entry marked (*) in each row indicates the best result.
For all methods, we run MCMC given all observations from the combined training-plus-validation dataset, and report likelihoods on the test set.
For single-site GGP and single-site GAR, we select hyperparameters values via grid search on the validation set.
For multi-site GAR, we set window size to $W=1$ to keep runtime affordable.
Tab.~\ref{tab:mass_data_quantitative_results_expanded} gives expanded results showing consistency across multiple MCMC chains.
}
\label{tab:mass_data_quantitative_results}
\label{tab:uk_data_quantitative_results}
\end{table}

\begin{figure}[!t]
\setlength{\tabcolsep}{0.1cm}
\centering
\begin{tabular}{c c}
	\scriptsize Melrose-Wakefield Hospital & \scriptsize Tufts Medical Center \\
    \includegraphics[width=0.49\textwidth]{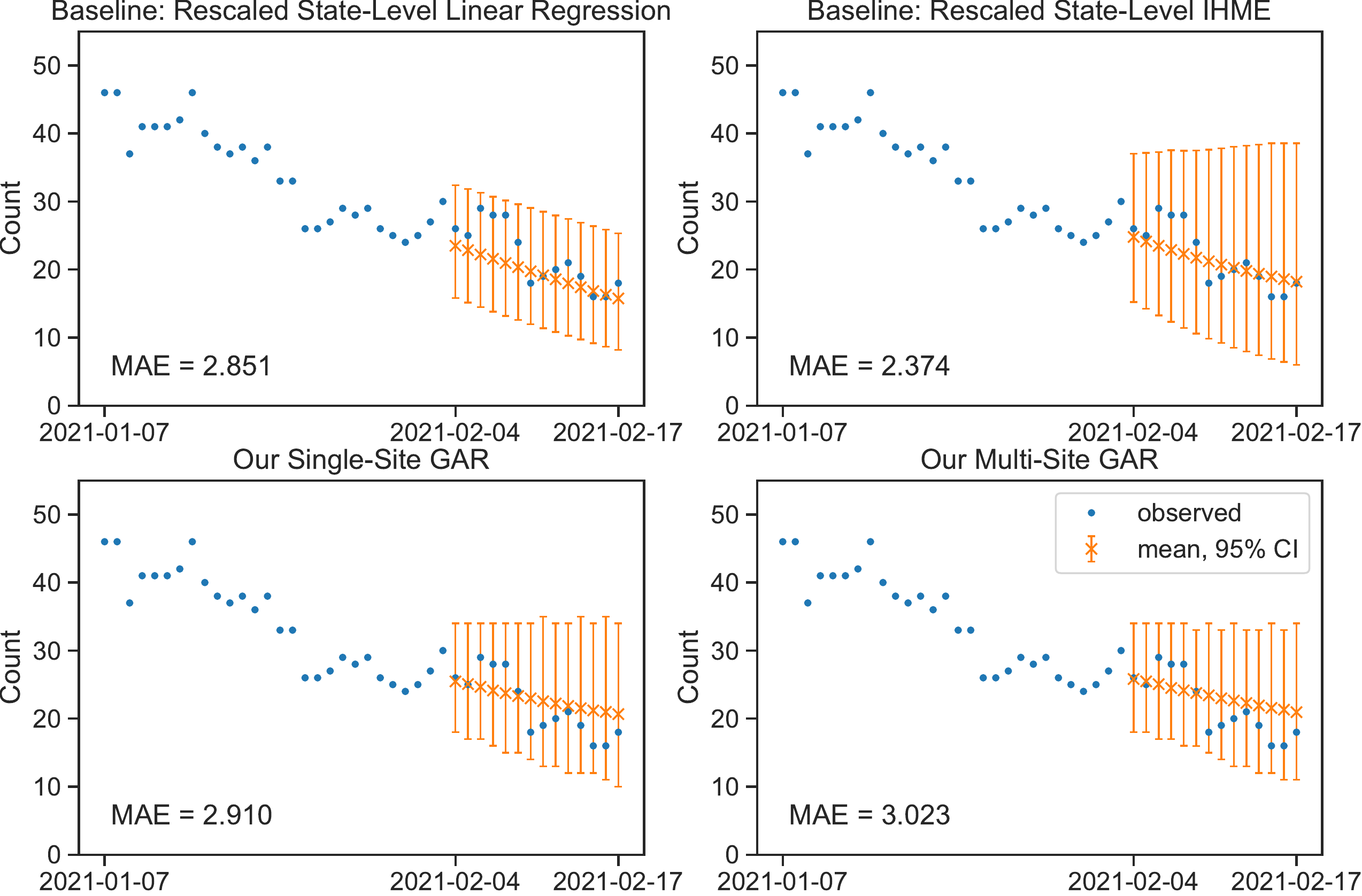}
	&
    \includegraphics[width=0.49\textwidth]{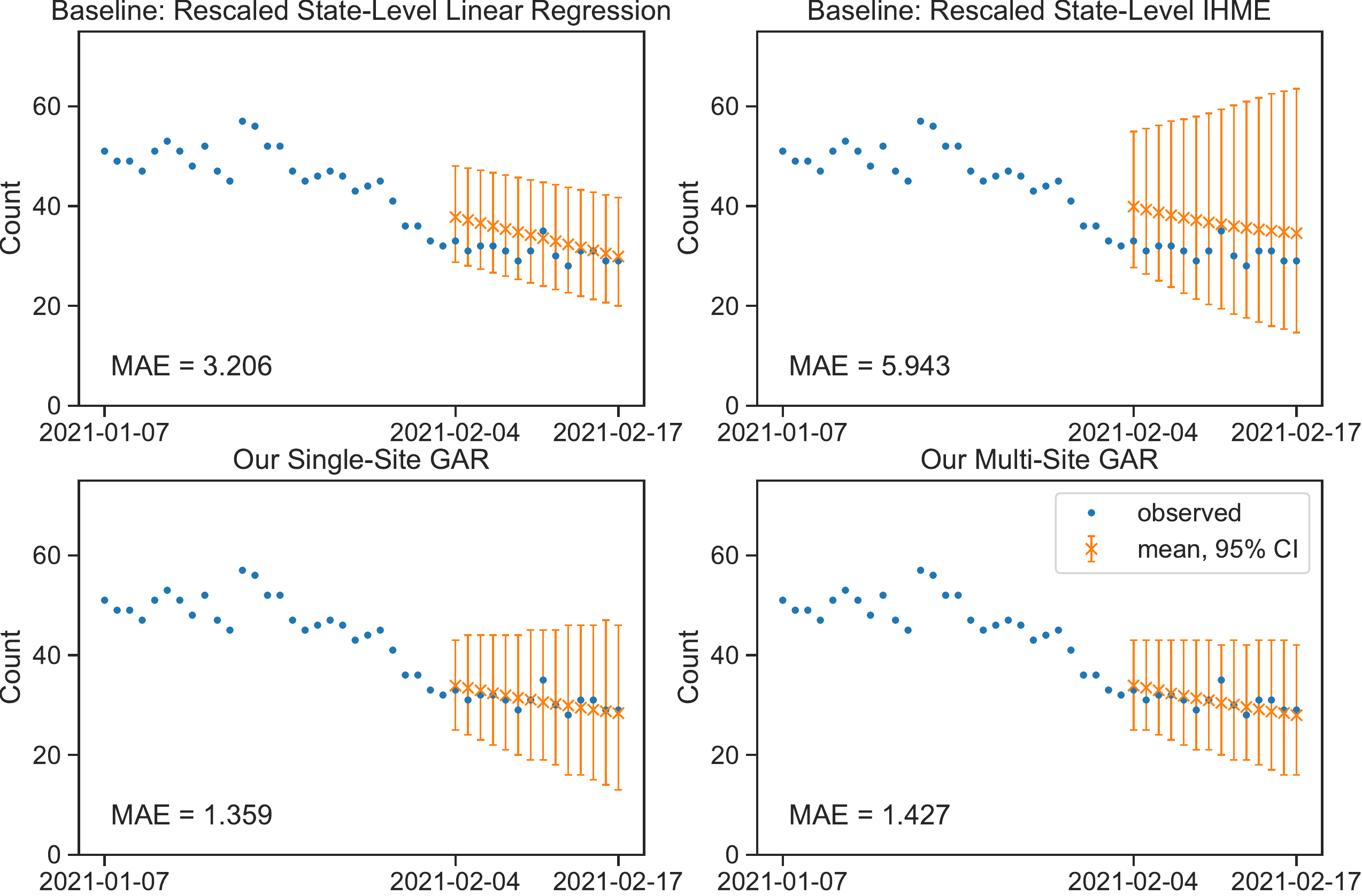}
\end{tabular}
\caption{
\textbf{Prospective evaluation of proposed models against baselines in use at two sites of a major hospital system in MA.}
All proposed GAR models deliver competitive mean absolute error and \emph{plausible} interval estimates where uncertainty increases slightly with time.
See expanded results for all sites in Fig.~\ref{fig:prospectiveEval_expanded}.
}
\label{fig:prospectiveEval_main}
\end{figure}

\textbf{Prospective validation: Methods}.
We performed 2-week-ahead forecasts starting on February 4, 2021 at 3 sites in Massachusetts - Melrose-Wakefield Hospital, Lowell General Hospital, and Tufts Medical Center - operated by a common healthcare system.
We compare our GAR models against two baselines \emph{currently in use} to help with 2-week-ahead planning at these sites.
The first baseline is a \textbf{rescaled state-level linear regression}, where state-level census counts are predicted with standard frequentist linear regression (unknown slope, unknown intercept) from the day indicator $t$. We use public data released by the state and fit to the previous 28 days of data.
The second baseline comes from \textbf{rescaled state-level IHME forecasts}, which uses the daily mean and 95\% CI of hospitalization forecasts for the state of MA made public by the Institute for Health Metrics and Evaluation (IHME).
Both baselines involve rescaling state-wide forecasts to site-specific levels. We first fit a linear regression to predict the \emph{fraction} $s^h_t$ of the total state-level volume at each site $h$ on each future day $t$, based on the previous 28 days of observed fractional volume.
Then for each day, we multiply the predicted fractional volume at each site of interest by the predicted state-level count.
We further estimate 95\% CIs of this fraction and use these to estimate site-level 95\% CIs.

\textbf{Prospective validation: Results}.
Fig.~\ref{fig:prospectiveEval_main} shows qualitative performance and mean absolute error (MAE) metrics on two of the sites in our prospective evaluation (see all 3 sites in Fig.~\ref{fig:prospectiveEval_expanded}). 
All methods capture the essentially linear trend of the test period in all 3 sites.
In most cases, the rescaled IHME baseline appears overly uncertain (intervals are too wide), so rescaled linear regression seems to be the better baseline.
At two sites, there seems to be little difference (MAE difference within 1.0) between our GAR methods and rescaled linear regression.
However, at one site (Tufts Medical) our methods appear noticeably better than the linear regression baseline: MAE improvement in these counts is larger than 1.75 patients per day. Furthermore, at all sites our methods produce the most reasonable uncertainty intervals, which sensibly \emph{grow slightly} further into the future.

\section{Discussion}

\textbf{Advantages.} Our approach is \emph{simple} yet \emph{effective}: we find that order-1 autoregressive latent variable models deliver reasonable predictions when we share latent dynamics parameters $\alpha$ across sites and use generalized Poisson likelihoods with flexible dispersion.
Our approach may be \emph{portable} to health systems around the world, as it relies on easy-to-collect count data and does not require much outside of a site leader's control (such as region-level data or expensive computing resources).
Our approach is \emph{robust} to realistic scenarios where some counts may be missing or corrupted: our fully-probabilistic model allows us to properly calibrate our uncertainty in these cases.
Our approach is \emph{extensible}, relying upon widely-used probabilistic programming toolkits ~\citep{salvatierProbabilisticProgrammingPython2016}.

\textbf{Limitations.}
The primary limitation of this work is that that our forecasts are purely based on statistical patterns in past census counts within the site(s) of interest.
Our generalized autoregressive approach may poorly predict in cases where underlying dynamics change (e.g. if the testing period sees higher vaccination rates) or in cases where the local trend is far from linear (hence our initial interest in GPs; we plan to do follow-up work to explore GPs with appropriately learned non-linear mean functions).
Additional data, especially leading indicators such as surveillance tests in the region or levels of community mobility and interaction, could improve forecasts especially when there is a \emph{regime change} in disease dynamics (e.g., from the summer lull to the late fall 2020 surge).
Other helpful data might include community demographics and density as well as referral patterns (e.g., from nursing homes).
We assume that hospitals in the same geographic region will have similar trends, which may miss how nearby hospitals serve different populations and thus see diverse trends. We do not account for how hospitals may \emph{interact} (e.g., transferring patients) to balance capacity.

While there is much more to be done, we hope this study raises interest in the single-site forecasting problem and offers a step forward for improving decision-making for local leaders.



\bibliography{c19.bib}

\appendix
\counterwithin{table}{section}
\setcounter{table}{0}
\counterwithin{figure}{section}
\setcounter{figure}{0}

\section{Datasets}
\label{supp:datasets}

\paragraph{Massachusetts data.}
We selected 8 hospital sites in Suffolk County, Massachusetts, USA targeted at adult patients with the largest volume of cases (all other sites had fewer than 50 patients per day). We used public data sourced from the Department of Public Health of the state of Massachusetts\footnote{\url{https://www.mass.gov/info-details/covid-19-response-reporting}}, which provided the total count of hospitalized patients with either suspected or confirmed cases of COVID-19, in both the general hospital or intensive care unit.

\begin{figure}[!t]
    \centering
    \includegraphics[width=0.7\textwidth]{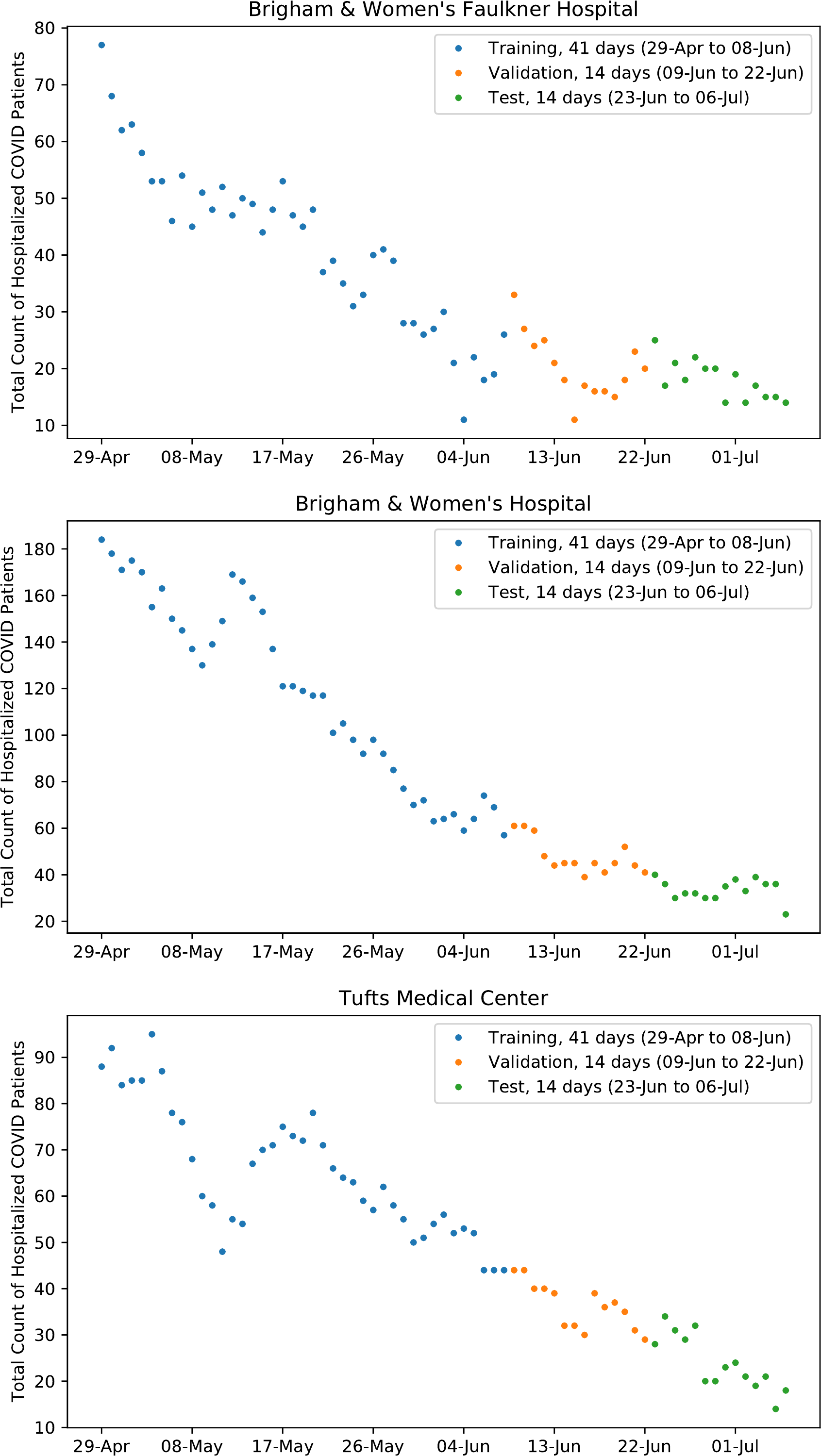}
    \caption{\textbf{Raw data from 3 representative hospital sites in our Massachusetts dataset.} We show the training, validation, and test splits used in experiments.}
    \label{fig:raw_data}
\end{figure}

The observed counts for 3 representative sites (out of the 8 total) for our selected study period are shown in Fig.~\ref{fig:raw_data}. In this data, we see a general decline across all sites, as expected as the data is from a transitory phase of the pandemic in Massachusetts from the spring surge leading into the summer lull.

\paragraph{UK data.}
We selected 10 hospital sites in England, United Kingdom with consistent data availability and the largest volume of cases. We used public data sourced from the National Health Service of the UK\footnote{\url{https://www.england.nhs.uk/statistics/statistical-work-areas/covid-19-hospital-activity/}}, which provided the count of beds occupied by COVID-19 patients on each day at each hospital.
We excluded sites whose datasets contained either zeros or counts that had a difference of greater than 50 from the previous 4 days and next 4 days, assuming those unrealistic values were inaccurate.

\section{Results on Retrospective Tasks}

For our retrospective evaluation, we selected a study period of April 29, 2020 to July 6, 2020.
Both MA and UK datasets had data available in this period.
To evaluate our forecasts, we chose to set our future duration to $F=14$ days (2 weeks). We selected this because 2 weeks into the future is a sensible planning horizon that allows some actionable logistics (expanding available beds, hiring traveling nursing staff) while remaining potentially predictable (forecasting the dynamics of hospitalizations caused by this disease further than 2 weeks from the present is a far more challenging problem with a low likelihood of success without incorporating other leading indicators such as testing rates and community activity levels).
We set aside the last 14 days of data as the test set (used only for computing heldout likelihoods), and then treated the 14 days before that as the validation set (used for hyperparameter selection). The remaining $T=41$ days served as our training set.

\subsection{Comparison of Likelihoods: Generalized vs. Standard Poisson}

Our first experiment sought to justify which likelihood model between the standard and generalized Poisson best suits our observed count data. While the standard Poisson distribution is more commonly used, the generalized Poisson offers more flexibility in dispersion.

\begin{figure}
    \centering
    \includegraphics[width=0.9\textwidth]{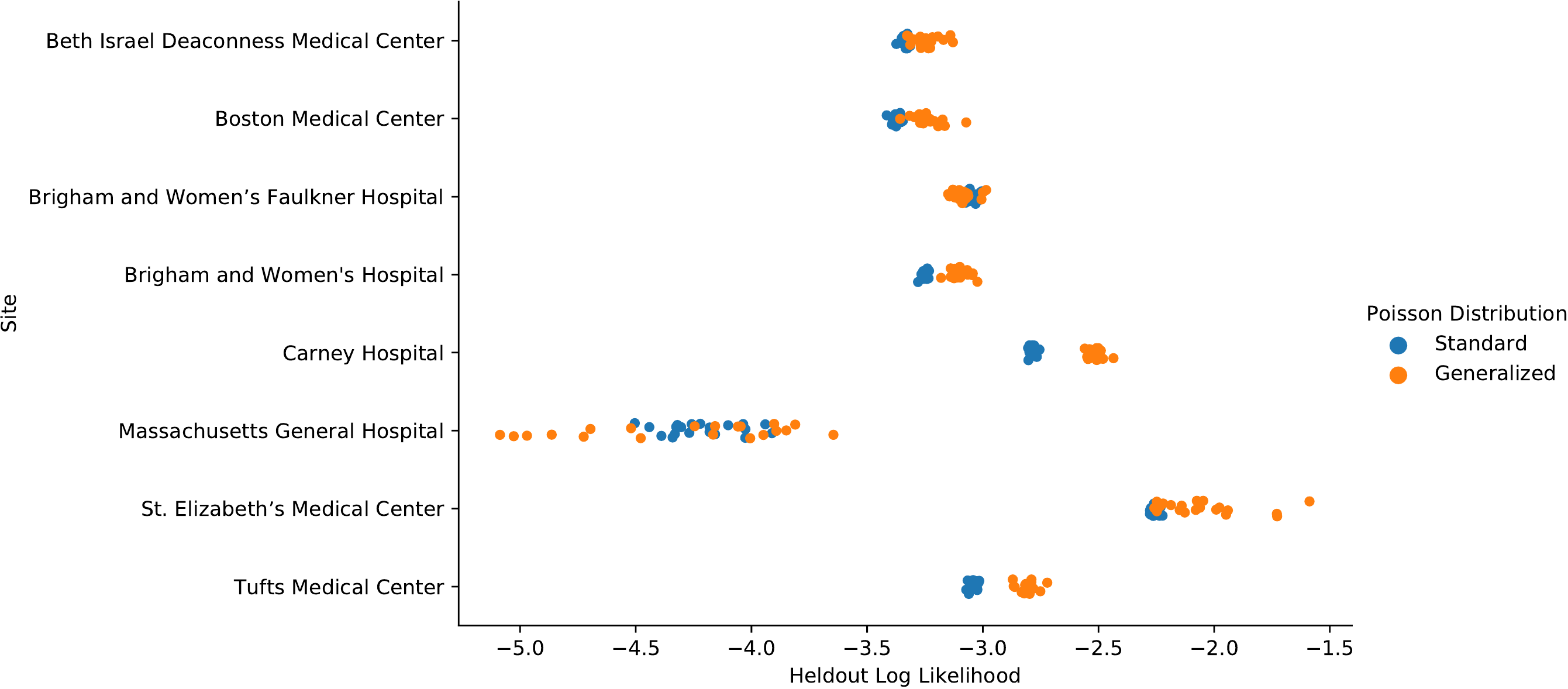}
    \caption{\textbf{Comparison of standard and generalized Poisson likelihoods.} The generalized Poisson outperforms the standard Poisson in general, with clear improvement in six sites and indistinguishable performance at two sites (Faulkner and Mass. General). For each site, there are 20 orange points (generalized Poisson) and 20 blue points (standard Poisson), showing estimates of the count-normalized log likelihood on the validation set after training the single-site GAR model with $W=1$ (simplest model). The 20 visualized points for each model come from 2 separate MCMC chains of 5000 samples each, divided into 10 groups of 500, to better indicate any mixing problems within chains.}
\label{fig:poisson_vs_genpoisson}
\end{figure}

We trained the single-site generalized auto-regressive (GAR) model for all 8 sites from Massachusetts with $W=1$ using both the standard Poisson and the generalized Poisson. Figure~\ref{fig:poisson_vs_genpoisson} shows a plot of multiple estimates of the heldout log likelihoods on the validation set under both likelihoods. These multiple estimates (derived from several Monte Carlo estimates from different MCMC chains) help us understand when differences are due to the underlying model and not just luck of the draw.

The primary takeaway is that we see greater heldout likelihoods when we use the generalized Poisson (clear improvement in 6 out of the 8 sites and indistinguishable performance in the other 2). Recall that using the standard Poisson likelihood is equivalent to using the generalized Poisson with $\lambda$ fixed at 0 rather than being a learned parameter. While the two models learn similar $\beta$ coefficients, the learned $\lambda$ parameter for the generalized Poisson concentrates well below zero for all sites, 
indicating the model is able to adjust for the underdispersion in the data. As a result, we don't overestimate the variance in the data and have greater certainty in our forecasts.

\subsection{Comparison of Latent Function Models: AR vs GP}

Next, we sought to understand which of our latent function models---the autoregressive (GAR) or Gaussian process (GGP)---offered the best fit to our data. We use a generalized Poisson likelihood for all these experiments.

\paragraph{Experimental Setup.} For both single-site models (GAR and GGP), we ran a grid search over hyperparameters. For each hyperparameter configuration, we condition on the training set when sampling from the posterior and then evaluate likelihoods on the validation set. We then combined the training and validation sets together using the hyperparameters with the best heldout performance, and evaluated the result on the test set.

For the single-site GAR model, we searched the following set of window sizes $W$: [1, 2, 5, 7, 10, 14]. For the single-site GGP model, we searched the following set of time-scale prior means $\mu_l$: [0, 5, 10, 15, 20, 25, 30, 35, 40, 45, 50].

We further evaluated the multi-site GAR model. Here, we fixed a window size of $W=1$, avoiding an expensive grid search. We found that performance was very similar between $W=1$ and $W=2$, especially averaged across sites, while $W=1$ was substantially faster to train (at least 5 times faster).

\begin{table}[!h]
\scriptsize
\begin{center}
\begin{tabular}{|p{3.5cm}| p{2.2cm}| p{2.2cm}| p{2.2cm}|} 
 \hline
 \textbf{Hospital Site (MA, USA)}
 & \textbf{Single-Site} \textbf{GGP}
 & \textbf{Single-Site} \textbf{GAR}
 & \textbf{Multi-Site} \textbf{GAR} \\
 \hline
 Beth Israel Deaconness
 	& \makecell[l]{$-3.153 \pm 0.005$ \\ $-3.148 \pm 0.003$} ~
 	& \makecell[l]{$-3.121 \pm 0.017$ \\ $-3.127 \pm 0.020$} ~
 	& \makecell[l]{$-3.010 \pm 0.011$ \\ $-3.011 \pm 0.012$} * \\
 \hline
 Boston Medical Center
 	& \makecell[l]{$-3.376 \pm 0.002$ \\ $-3.379 \pm 0.004$} ~
 	& \makecell[l]{$-3.454 \pm 0.025$ \\ $-3.406 \pm 0.021$} ~ 
 	& \makecell[l]{$-3.296 \pm 0.014$ \\ $-3.295 \pm 0.015$} * \\
 \hline
Brigham \& Women's Faulkner 
	& \makecell[l]{$-2.719 \pm 0.002$ \\ $-2.719 \pm 0.003$} ~
	& \makecell[l]{$-2.684 \pm 0.010$ \\ $-2.671 \pm 0.014$} ~
	& \makecell[l]{$-2.601 \pm 0.008$ \\ $-2.583 \pm 0.006$} * \\
 \hline
Brigham \& Women’s Hospital
	& \makecell[l]{$-3.118 \pm 0.001$ \\ $-3.114 \pm 0.001$} ~
	& \makecell[l]{$-2.978 \pm 0.009$ \\ $-2.948 \pm 0.012$} *
	& \makecell[l]{$-3.044 \pm 0.016$ \\ $-3.069 \pm 0.027$} ~ \\
 \hline
 Carney Hospital
 	& \makecell[l]{$-3.072 \pm 0.008$ \\ $-3.073 \pm 0.007$} ~
 	& \makecell[l]{$-2.721 \pm 0.105$ \\ $-2.551 \pm 0.080$} ~
 	& \makecell[l]{$-2.564 \pm 0.090$ \\ $-2.619 \pm 0.079$} * \\
 \hline
Massachusetts General
	& \makecell[l]{$-3.777 \pm 0.020$ \\ $-3.775 \pm 0.019$} ~
	& \makecell[l]{$-3.468 \pm 0.018$ \\ $-3.463 \pm 0.013$} *
	& \makecell[l]{$-3.450 \pm 0.017$ \\ $-3.490 \pm 0.025$} * \\
 \hline
St. Elizabeth’s
	& \makecell[l]{$-1.980 \pm 0.007$ \\ $-1.983 \pm 0.010$} ~
	& \makecell[l]{$-1.312 \pm 0.120$ \\ $-1.129 \pm 0.122$} *
	& \makecell[l]{$-1.598 \pm 0.007$ \\ $-1.609 \pm 0.002$} ~ \\
 \hline
 Tufts Medical Center
 	& \makecell[l]{$-2.937 \pm 0.005$ \\ $-2.936 \pm 0.004$} ~
 	& \makecell[l]{$-2.822 \pm 0.010$ \\ $-2.821 \pm 0.016$} ~
 	& \makecell[l]{$-2.728 \pm 0.007$ \\ $-2.750 \pm 0.007$} * \\
 \hline
\end{tabular}
\begin{tabular}{|p{3.5cm}|p{2.2cm}|p{2.2cm}|p{2.2cm}|} 
\hline
 \textbf{Hospital Site (UK)}
 & \textbf{Single-Site} \textbf{GGP}
 & \textbf{Single-Site} \textbf{GAR}
 & \textbf{Multi-Site} \textbf{GAR} \\
 \hline
 Barts Health
 	& \makecell[l]{$-4.730 \pm 0.058$ \\ $-4.706 \pm 0.043$} *
 	& \makecell[l]{$-6.727 \pm 0.221$ \\ $-6.758 \pm 0.118$} ~
 	& \makecell[l]{$-4.842 \pm 0.078$ \\ $-4.852 \pm 0.055$} ~ \\
 \hline
 Chelsea \& Westminster 
 	& \makecell[l]{$-2.937 \pm 0.009$ \\ $-2.942 \pm 0.013$} *
 	& \makecell[l]{$-3.141 \pm 0.030$ \\ $-3.183 \pm 0.023$} ~
 	& \makecell[l]{$-5.929 \pm 0.086$ \\ $-5.880 \pm 0.106$} ~ \\
 \hline
 Imperial College Healthcare
 	& \makecell[l]{$-4.017 \pm 0.019$ \\ $-4.039 \pm 0.049$} ~
 	& \makecell[l]{$-3.063 \pm 0.178$ \\ $-3.116 \pm 0.145$} *
 	& \makecell[l]{$-3.976 \pm 0.110$ \\ $-4.100 \pm 0.107$} ~ \\
 \hline
 King's College Hospital
 	& \makecell[l]{$-3.012 \pm 0.003$ \\ $-3.017 \pm 0.003$} ~
 	& \makecell[l]{$-2.749 \pm 0.020$ \\ $-2.735 \pm 0.014$} ~
 	& \makecell[l]{$-2.695 \pm 0.005$ \\ $-2.695 \pm 0.007$} *\\
 \hline
 London North West University
 	& \makecell[l]{$-2.829 \pm 0.008$ \\ $-2.807 \pm 0.017$} ~
 	& \makecell[l]{$-2.488 \pm 0.014$ \\ $-2.506 \pm 0.017$} *
 	& \makecell[l]{$-2.572 \pm 0.009$ \\ $-2.560 \pm 0.007$} ~ \\
 \hline
 Manchester University
 	& \makecell[l]{$-12.335 \pm 0.647$ \\ $-11.275 \pm 0.807$} ~
 	& \makecell[l]{$-12.735 \pm 0.937$ \\ $-13.028 \pm 0.924$} ~
 	& \makecell[l]{$-6.152 \pm 0.167$ \\ $-6.021 \pm 0.241$} * \\
 \hline
 North Middlesex University
 	& \makecell[l]{$-1.593 \pm 0.008$ \\ $-1.588 \pm 0.006$} *
 	& \makecell[l]{$-1.677 \pm 0.022$ \\ $-1.656 \pm 0.015$} ~
 	& \makecell[l]{$-1.703 \pm 0.008$ \\ $-1.708 \pm 0.012$} ~ \\
 \hline
 Nottingham University
 	& \makecell[l]{$-3.443 \pm 0.018$ \\ $-3.472 \pm 0.018$} ~
 	& \makecell[l]{$-3.324 \pm 0.066$ \\ $-3.275 \pm 0.047$} ~
 	& \makecell[l]{$-2.950 \pm 0.032$ \\ $-2.986 \pm 0.026$} * \\
 \hline
 University Hospitals Birmingham
 	& \makecell[l]{$-4.385 \pm 0.005$ \\ $-4.371 \pm 0.010$} *
 	& \makecell[l]{$-4.571 \pm 0.019$ \\ $-4.548 \pm 0.033$} ~
 	& \makecell[l]{$-4.583 \pm 0.031$ \\ $-4.620 \pm 0.023$} ~\\
 \hline
 Univ. Hospitals of North Midlands
 	& \makecell[l]{$-3.188 \pm 0.002$ \\ $-3.187 \pm 0.003$} ~
 	& \makecell[l]{$-2.953 \pm 0.033$ \\ $-2.993 \pm 0.028$} ~
 	& \makecell[l]{$-2.829 \pm 0.016$ \\ $-2.841 \pm 0.016$} * \\
 \hline
\end{tabular} 
\end{center}

\caption{\textbf{Heldout likelihood evaluation across 8 sites in MA (top) and 10 sites in UK (bottom).} We report test-set log likelihood (normalized by total test days). The entry marked (*) in each row indicates the best result.
For each model, we report results from 2 separate MCMC chains to assess reliability (large differences across chains indicate convergence problems).
Within each chain of 5000 samples, we gather 10 groups of 500 samples, and report the mean $\pm$ SEM to communicate uncertainty.
For all methods, we run MCMC given all observations from the combined training-plus-validation dataset, and report likelihoods on the test set.
For single-site GGP and single-site GAR, we select hyperparameters values via grid search on the validation set.
For multi-site GAR, we set the window size to $W=1$ to keep runtime affordable.
}
\label{tab:mass_data_quantitative_results_expanded}
\label{tab:uk_data_quantitative_results_expanded}
\end{table}

\begin{figure}[!h]
\setlength{\tabcolsep}{0.01cm}
\begin{tabular}{c}
    \centering
    \includegraphics[width=0.95\textwidth]{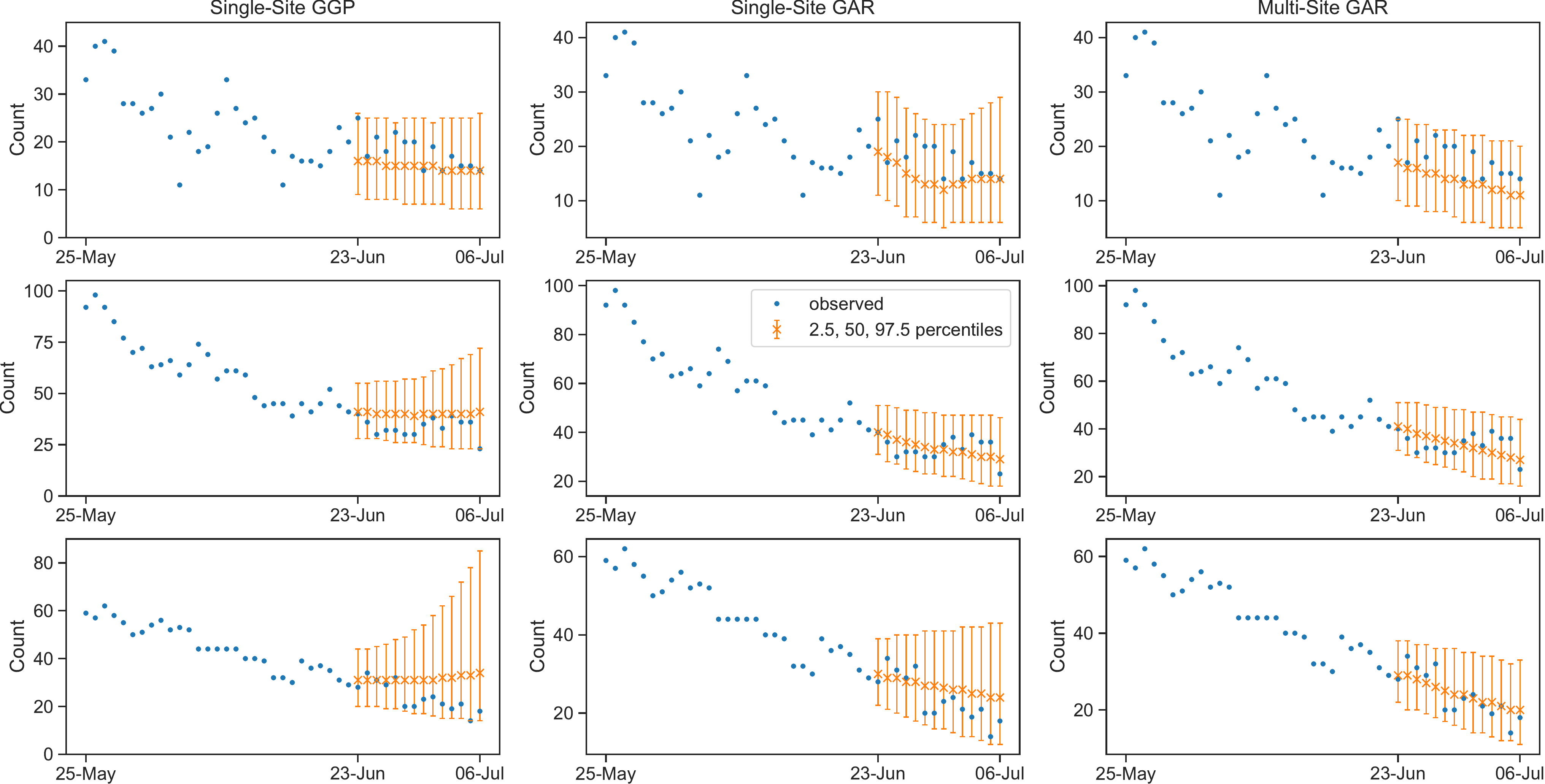}
\end{tabular}
\caption{
    \textbf{Qualitative forecasts for 3 representative hospital sites selected from the 8 available sites in our MA dataset.}
    From top to bottom, the 3 sites are: Brigham \& Women's Hospital, Brigham \& Women's Faulkner Hospital, and Tufts Medical Center.
    Each column shows one of our methods of interest: single-site GGP, single-site GAR, and multi-site GAR.
    We show the true observed values and the 2.5\textsuperscript{th}, 50\textsuperscript{th}, and 97.5\textsuperscript{th} percentiles of the sampled distribution on each day (S=5000 total samples).
    In general, the GGP (left) sometimes learns very wide predictive intervals, while the multi-site GAR (right) delivers better calibrated intervals.
}
\label{fig:mass_data_qualitative_forecasts}
\end{figure}

\paragraph{Qualitative Results.} 
Figure~\ref{fig:mass_data_qualitative_forecasts} shows qualitative forecasts for 3 representative sites produced by the three models of interest (single-site GGP, single-site GAR, and multi-site GAR). 

We find in general that the GAR model is more reliable. Although the GGP can sometimes achieve better performance, it is difficult to find its ideal parameters. Because the sequences of census counts during our study period are relatively linear, likely a GP with a learnable linear mean function would have done better. Under our chosen model with a non-zero constant mean function, large time-scales are required to achieve better predictive performance. We chose a prior distribution over time-scales with larger mean and small standard deviation in order to inductively bias the posterior towards larger time scales; with a larger prior standard deviation and more freedom, the posterior time-scale might instead become smaller (improving some training predictions at the cost of generalization).
The single-site GGP tends to be sensitive to the variability in the data over smaller time windows, causing the predictions curve away from the training data and prediction intervals to expand. On the other hand, the GAR learns the overall linear nature of the time series, leading to flatter predictions and intervals that expand less. The multi-site GAR achieves even greater certainty with more training data and less sensitivity to variability in the data.

\paragraph{Quantitative Results.}
Table \ref{tab:mass_data_quantitative_results_expanded} gives the quantitative results for the Massachusetts data and UK data.
We show heldout log likelihood for all three models, indicating the mean $\pm$ the standard error of the mean (SEM) for multiple MCMC chains (to help diagnose any convergence issues).

\textbf{MA Results.}
In the experiments on Massachusetts data, the single-site GAR performed better than the single-site GGP, and the multi-site GAR performed better than the single-site GAR.
In 7 out of 8 sites, the single-site GAR achieved better performance than the single-site GGP. The multi-site GAR achieved better performance than the single-site GAR in 4 out of 8 sites, and similar performance in 3 sites. In all 8 sites, the multi-site GAR outperformed the single-site GGP. We conclude that the multi-site GAR is the best model for our data. While it isn't consistently better than the single-site GAR, it offers a much faster forecasting model because it requires us to train only one set of parameters for the entire set of sites, rather than one set of parameters for each site.

\textbf{UK Results.}
In general, we had greater success with the Massachusetts data than the UK data. We did not see the multi-site GAR generally outperform the other models on the UK data because there was less similarity between hospital sites. While the hospital sites in the Massachusetts dataset are all in the same Massachusetts county (within about 30 miles), the hospital sites in the UK dataset are from a much larger region (within about 300 miles). Additionally, while the Massachusetts datasets are relatively smooth, the UK datasets have a lot more jumps and irregularities. For the sites that have smoother count series (like King's College Hospital, Nottingham University Hospitals, and University Hospitals of North Midlands), we see the same pattern as the Massachusetts data, where the GAR outperforms the GGP, and the multi-site model outperforms the single-site model. On the other hand, we don't see the same trends in the results for sites with more jumps and irregularities (like Chelsea \& Westminster Hospital and Imperial College Healthcare). The gains from GGP to GAR and from single-site to multi-site come from tighter predictions/less uncertainty, but the datasets that are less smooth don’t benefit from that.

\section{Results on Prospective Tasks}

\begin{figure}[!h]
\setlength{\tabcolsep}{0.3cm}
\begin{tabular}{m{1.5cm} m{8cm}}
Lowell General Hospital &     \includegraphics[width=0.7\textwidth]{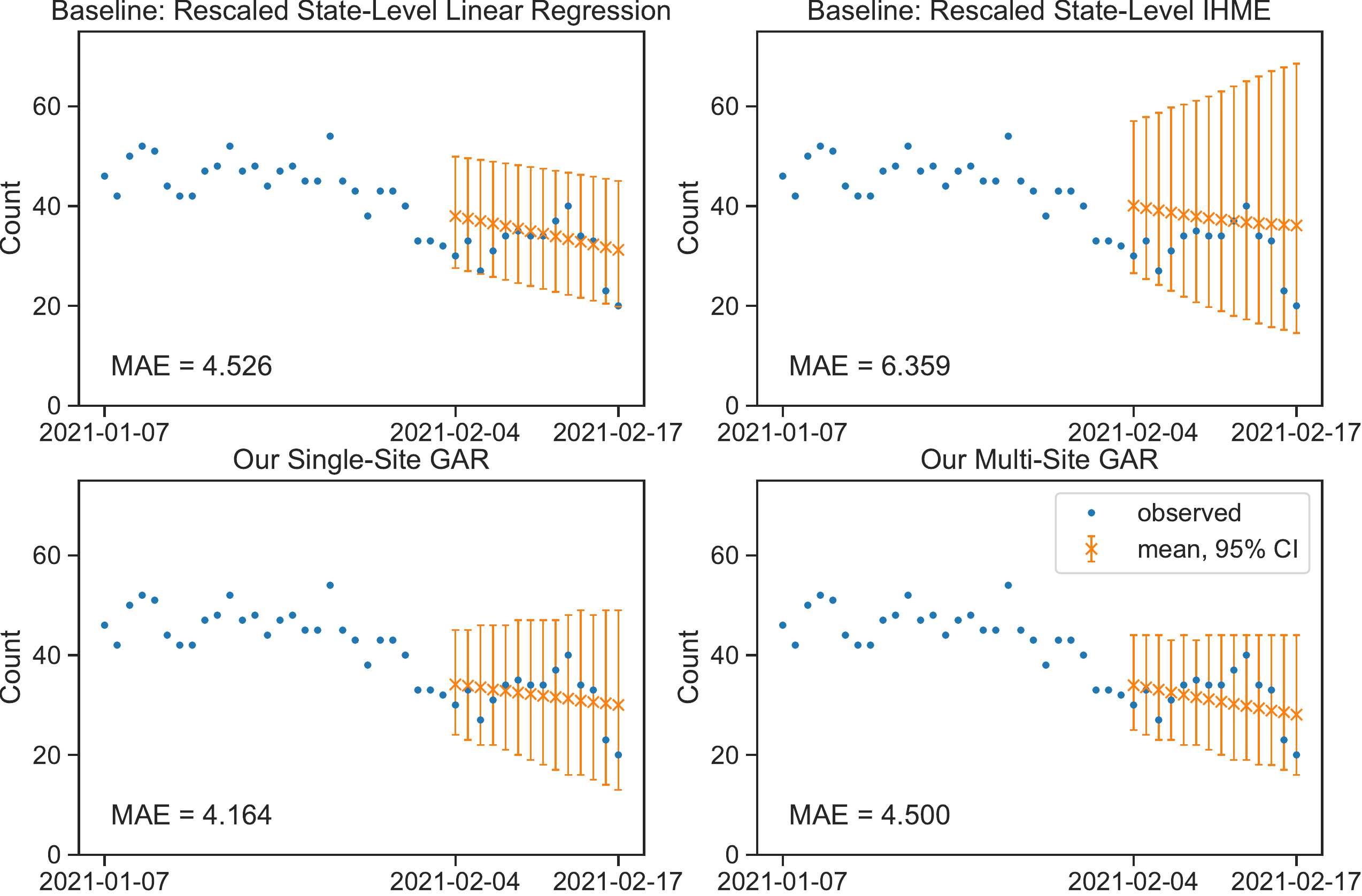}
	\\ ~ \\~ \\ 
Melrose-Wakefield Hospital &    \includegraphics[width=0.7\textwidth]{figures/MWH.pdf}
	\\ ~ \\~ \\ 
Tufts Medical Center & 
    \includegraphics[width=0.7\textwidth]{figures/TMC.pdf}
\end{tabular}
\caption{
\textbf{Prospective evaluation of proposed models against baselines in use at all 3 hospital sites of the major hospital system we studied.}
All proposed GAR models deliver competitive mean absolute error and \emph{plausible} interval estimates where uncertainty increases slightly with time.
}
\label{fig:prospectiveEval_expanded}
\end{figure}

We show in Fig.~\ref{fig:prospectiveEval_expanded} an expanded set of qualitative visual forecasts from our prospective evaluation on all 3 sites of the hospital system of interest in Massachusetts. The figure also indicates the mean absolute error (MAE) of each method.

A few key conclusions of our prospective evaluation are noticeable from this figure:

\begin{itemize}
\item All methods capture the essentially linear trend of the data in the testing period reasonably well.
\item Between the two baselines (rescaled linear regression, rescaled IHME forecasts), we see consistently larger uncertainty intervals for the IHME forecasts. Rescaled linear regression appears to be the stronger baseline in terms of MAE: improvements of over 2.0 at two sites (LGH and TMC), and a difference under 0.5 at the third site (Melrose-Wakefield). 
\item The single-site GAR and multi-site GAR deliver similar qualitative trends and prediction accuracy. The multi-site GAR has slightly worse MAE, but this is likely insignificant (its MAE is within 0.5 of the single-site model, so the daily difference in counts is less than one patient). However, the multi-site GAR gives uncertainty intervals that are narrower while still fully capturing the true values. (Note that MAE is calculated using only the mean prediction values and doesn't account for uncertainty, which is a key goal of our forecasts.)
\item Our single-site and multi-site GAR models appear to either match or outperform our baseline models on all three sites. We emphasize that our models' uncertainty intervals sensibly grow over time (unlike the frequentist linear regression baseline), accounting for larger uncertainty further in the future. 
\end{itemize}

\clearpage

\section{Method Details: Likelihoods for Count Data}
\label{sec:background-count-likelihoods}

In this section, we review possible distributions for modeling observed data $y$ which represents \emph{counts} or non-negative integers: $y \in \{0, 1, 2, \ldots \}$.

\paragraph{Poisson Likelihood.}
Recall the Poisson distribution over non-negative integer random variable $Y$ has probability mass function:
\begin{align}
    p(Y = y \mid \theta) = \frac{1}{y!} \theta^y e^{-\theta},
    \qquad y \in \{0, 1, 2, \ldots \},
\end{align}
where the parameter $\theta > 0$ controls both the mean and the variance of $Y$:
\begin{align}
    \mathbb{E}[Y] = \theta,
    \quad \text{Var}[Y] = \theta.
\end{align}

\paragraph{Generalized Poisson Likelihood.}
The generalized Poisson distribution \citep{consulGeneralizedPoissonRegression1992} is an extension of the standard Poisson. Unlike the standard Poisson distribution, which has a single parameter determine both mean and variance and thus assumes equidispersion (the mean is the same as the variance), the generalized Poisson has a separate dispersion parameter that allows us to model count data that is either underdispersed (variance is less than the mean value) or overdispersed (variance is greater than the mean value).

A generalized Poisson random variable $Y$ has the following probability mass function:
\begin{align}
    p(Y = y \mid \theta, \lambda) = \frac{1}{y!} \theta (\theta + \lambda y)^{y-1} e^{-\theta - \lambda y},
    \qquad y \in \{0, 1, 2, \ldots \}, \label{eq:gen-poisson-pmf}
\end{align}
where $\theta > 0$ and $\max(-1, -\frac{\theta}{4}) \leq \lambda \leq 1$. The mean and variance are given by
\begin{align}
    \mathbb{E}[Y] = \frac{\theta}{1 - \lambda},
    \quad \text{Var}[Y] = \frac{\theta}{(1 - \lambda)^3}.
\end{align}
When $\lambda = 0$, the generalized Poisson reduces to the standard Poisson with mean $\theta$. When $\lambda < 0$, the model has underdispersion; when $\lambda > 0$, the model has overdispersion.

\paragraph{Priors on likelihood dispersion parameter $\lambda$.}
For the generalized Poisson, we make the a priori assumption that the count data has symmetric chances of both under- and over-dispersion. We thus set the prior on the dispersion parameter $\lambda$ to be
\begin{align}
    \lambda \sim \text{TruncatedNormal}(0, 0.3, \text{lower}=-1, \text{upper}=1).
\end{align}

\paragraph{Sampling from the generalized Poisson.}
We generate random samples from a generalized Poisson using the Inversion Algorithm~\citep{famoyeGeneralizedPoissonRandom1997}, whereby a random sample is drawn from the uniform distribution and then plugged into the inverse of the cumulative distribution function. The inverse CDF of the generalized Poisson distribution is available in closed-form, making this possible.

\section{Method Details: Single-Site GAR}
\label{sec:methods-single-site-GAR}

First, we consider a latent autoregressive process for count data (which we call the generalized autoregressive model or ``GAR'').
This model is ``generalized'' in the same sense as generalized linear models; that is, we may explore multiple possible likelihoods for count data (standard Poisson and generalized Poisson), rather than simply assuming the data is normally distributed.

Our generalized autoregressive model is an instance of the generative model $p(f_{1:T}, y_{1:T})$ in Eq.~\eqref{eq:latent-variable-time-series} with an order-$W$ autoregressive process for the latent-generating distribution over $f_{1:T}$
\begin{align}
    p_{\alpha}(f_{1:T}) = \prod_{t=1}^T p_{\alpha}( f_t \mid f_{(t-W_t):(t-1)} ), \quad W_t = \text{min}(t-1,W)
\end{align}
where the recent window size $W$ is a hyperparameter.

For most timesteps $t > W$, we generate latent value $f_t$ using the fully-available window of $W$ previous values $f_{t-W:t-1}$:
\begin{align}
	p_{\alpha}(f_t \mid f_{(t-W):(t-1)}) = \text{NormalPDF}(\beta_0 + \sum_{\tau = 1}^{W} \beta_{\tau} f_{t-\tau}, \sigma^2), ~ t \in W{+}1, W{+}2, \ldots
\end{align}
The latent-generating parameters $\alpha$ for the GAR are $\alpha = \{\beta, \sigma \}$, with coefficient vector $\beta = [\beta_0, \beta_1, \ldots \beta_W]$ and standard deviation $\sigma > 0$.

For the first $W$ timesteps, the full window is not available, and so we regress only on the available previous values:
\begin{align}
	p_{\alpha}(f_t \mid f_{1:(t-1)}) = \text{NormalPDF}(\beta_0 + \sum_{\tau = 1}^{t-1} \beta_{\tau} f_{t-\tau}, \sigma^2), ~ t \in 1, 2, \ldots W
\end{align}
The very first value $t=1$ is generated with mean $\beta_0$.

A simple case of this model occurs when $W=1$, yielding the order-1 autoregressive process, which generates each timestep's latent value using a regression on the previous value $f_{t-1}$:
\begin{align}
	f_t \mid f_{t-1} \sim \text{Normal}(
	    \beta_0 + \beta_{1} f_{t-1},
	    \sigma^2)
\end{align}

\paragraph{Priors on latent-generating parameters $\alpha$.}
For the GAR, our latent-generating parameters are $\alpha = \{ \beta, \tau \}$.
We place Normal priors over all of the coefficients $\beta$, all centered at 0 except for the coefficient on the most recent timestep, whose prior is centered at 1 as we can assume that the future is like the recent past:
\begin{gather}
    \beta_0 \sim \text{Normal}(0, 0.1), \\
    \beta_1 \sim \text{Normal}(1, 0.1), \\
    \beta_2 \sim \text{Normal}(0, 0.1), \\
    ... \\
    \beta_W \sim \text{Normal}(0, 0.1).
\end{gather}
We also assume that the standard deviation of the autoregressive process should be close to 0 (since we imagine the latent process as a clean signal of an overall trend which gives rise to noisier count observations). Thus, we set the prior on the standard deviation to be:
\begin{align}
	\sigma \sim \text{HalfNormal}(0.1).
\end{align}

\paragraph{Advantages and Limitations.}
Our GAR model is advantageous due to its simple form of dependency within the latent sequence $f_{1:T}$, which makes both learning and prediction fast. It is limited in the kinds of long-term dependencies it can capture, which motivates our next class of models.

\section{Method Details: Single-Site GGP}
\label{sec:methods-single-site-GGP}

We next consider a latent Gaussian process model, building on early work on Gaussian processes~ \citep{rasmussenGaussianProcessesMachine2006}, GP latent variable models~\citep{lawrenceGaussianProcessLatent2003}, GP state space models~\citep{damianouVariationalGaussianProcess2011}, and extensions of GPs to general likelihoods~\citep{chanGeneralizedGaussianProcess2011}. Our proposed GGP model uses GPs to capture correlations between latent values $f_{1:T}$. Again, this is an instance of the generative model $p(f_{1:T}, y_{1:T})$ in Eq.~\eqref{eq:latent-variable-time-series}, with a GP for the prior over $f_{1:T}$:
\begin{align}
    f_{1:T} \sim \text{GP}( m_{\alpha}(t), k_{\alpha}(t,t') )
\end{align} 
where we assume a constant mean and squared exponential covariance kernel:
\begin{align}
    m_{\alpha}(t) = c, \qquad
    k_{\alpha}(t, t') = a^2 \exp\left(-\frac{(t - t')^2}{2 \ell^2}\right)
\end{align}
Here, the set of latent-generating parameters $\alpha$ is $\alpha = \{c, a, \ell\}$, with $a > 0$ and $\ell > 0$.

\paragraph{Priors on latent-generating parameters $\alpha$.}
We assume that the constant mean function $c$ should be approximately equal to the mean of the set of observed $\log$ counts in our training set. We set the covariance amplitude $a$ to be close to 0 to keep noise in the latent sequence $f$ low.

Thus, we place the following priors over the Gaussian process parameters:
\begin{align}
    c &\sim \text{TruncatedNormal}(4, 2, \text{lower}=0) \\
    a &\sim \text{HalfNormal}(2) \\
    \ell &\sim \text{TruncatedNormal}(\mu_\ell, 2, \text{lower}=0)
\end{align}
Here, $\mu_\ell$ is a hyperparameter that controls the time-scale of dependencies. We set this via grid search on validation data.
We choose a small standard deviation for the prior on $\ell$ to keep the posterior distribution close to the hyperparameter values being evaluated.

\section{Method Details: Model Fitting and Evaluation}

Our goal is to make accurate predictions for \emph{future} data given historical counts. To assess each model's ability to do so, we partition the data into training and evaluation sets by time rather than by random selection. Given a time series of $T+F$ counts, we treat the first $T$ counts as \emph{observed} data, and try to determine our prediction quality for the last $F$ counts.

\subsection{Posterior analysis and forecasting}

Armed with our assumed generative model in Eq.~\eqref{eq:latent-variable-time-series} with either an AR or GP prior for the latent-sequence-generating distribution $p_{\alpha}(f_{1:T})$, our analysis goal is to take as input an observed time-series dataset of single site counts $y_{1:T}$ and 
use this to make useful predictions about future data $y_{(T+1):(T+F)}$. We achieve this in two steps. First, we need to draw samples of parameters $\alpha, \lambda$ and the past latent sequence $f_{1:T}$ from their posterior given the past counts $y_{1:T}$. Second, given these parameters and latents from the posterior we can either sample future counts or evaluate the likelihood of some given future counts, using the generative model defined by these parameters. 

\paragraph{Posterior sampling.}
Given an observed time-series dataset of single site counts, we wish to first sample from the posterior over parameters $\alpha$, $\lambda$, and latent values $f$: $p(\alpha, \lambda, f_{1:T} \mid y_{1:T})$. 
We use the No-U-Turn sampler \citep{hoffmanNoUTurnSamplerAdaptively2014} to perform Markov chain Monte Carlo approximation of this posterior.
Our NUTS sampler is implemented using the PyMC3 toolbox~\citep{salvatierProbabilisticProgrammingPython2016}.

\paragraph{Forecast sampling.} Given a single posterior sample indexed by $s$, with parameters $\alpha^s, \lambda^s$ and latents $f^s_{1:T}$, we can then use the generative model to draw a \emph{forecast} of latents $f$ and counts $y$ for the next $F$ days:
\begin{align}
    f^s_{T+\tau} &\sim  p_{\alpha^s}( f_{T+\tau} \mid f^s_{1:(T+\tau-1)} ), &\tau \in 1, \ldots F
    \\
    y^s_{T+\tau} &\sim p_{\lambda^s}( y_{T+\tau} \mid f^s_{T+\tau} ), &\tau \in 1, \ldots F
\end{align}
Naturally, the independence assumptions in the GAR model make the first equation above reduce to $p_{\alpha^s}( f_{T+\tau} \mid f^s_{(T+\tau-W):(T+\tau-1)} )$.

We typically draw a forecast for each of $S$ distinct samples, using $S=1000$ or more to be sure we're capturing the full distribution.
We can compute summary statistics of the empirical distribution over these $S$ samples such as mean or median values as well as lower and upper percentiles.

Using the previously described sampling and forecasting methods, we obtain $S$ samples,
\begin{align}
    \alpha^s, \lambda^s, \fpast^s, \ffuture^s \sim
    p( \alpha, \lambda, \fpast, \ffuture \mid \ypast ), \quad s \in 1, 2, \ldots S,
\end{align}
which we use to estimate the likelihood of future data given the past.

\paragraph{Computing Probability of Heldout Data.}
To score the model's performance, we evaluate the likelihood of heldout future counts:
\begin{align}
    p(\yfuture \mid \ypast) &= \int
        p( \yfuture, \ffuture, \lambda \mid \ypast) d\ffuture d\lambda \\
    &= \int p_{\lambda}(\yfuture \mid \ffuture) p(\ffuture , \lambda \mid \ypast) d\ffuture d\lambda
\end{align}
This is an expectation with respect to a posterior that conditions on $\ypast$.
Given any posterior sample from the full joint $p(\alpha, \lambda, \ffuture, \fpast \mid \ypast )$, we can always just drop $\alpha^s, \fpast^s$ to obtain sampled values $\lambda^s, \ffuture^s$ from the posterior needed for the expectation above: $p(\ffuture, \lambda \mid \ypast )$. Thus, given $S$ samples we can compute a Monte Carlo approximation of the integral above to estimate the \emph{log} probability of the heldout future counts:
\begin{align}
    \log p(\yfuture  \mid \ypast )
    &\approx \log \frac{1}{S} \sum_{s=1}^S p_{\lambda^s}( \yfuture \mid \ffuture^s )
    \\
    &\approx \log \frac{1}{S} \sum_{s=1}^S \prod_{\tau=1}^{F} p_{\lambda^s}(y_{T+\tau} \mid f_{T+\tau}^s )
    \label{eq:heldout_likelhood_montecarlo}
\end{align}
This uses our chosen count likelihood probability mass function $p_{\lambda}(\cdot)$, which is typically the generalized Poisson but could be either of the functions in Sec.~\ref{sec:background-count-likelihoods}.

\paragraph{Practical rescaling.} 
Across different real-world sites, the scale of the heldout log likelihood values computed by the above method is likely to differ substantially between large and small sites. To facilitate a more ``sensible'' scale, we recommend normalizing all likelihoods by the number of observations $F$. We thus compute $\frac{1}{F} \log p(y_{(T+1):(T+F)} \mid y_{1:T})$. We find that in practice this makes human interpretation of these values more sensible, as the value has a consistent scale on the order of -10 to -2 regardless of the length of the window $F$.

\section{Method Details: Multi-Site Models}
\label{sec:methods-multi-site}

We now aim to predict future census counts at $H$ different hospital sites simultaneously, given $T$ previous observations from each site.
The intuition is that all $H$ sites share common trends (i.e., whether they increase or decrease over short-term time scales), though there may be site-specific patterns in the observed data (e.g. one hospital may typically have many more patients than another).

We consider a multi-site generative model which ties the latent-sequence-generating parameters $\alpha$ across sites, but allows count-generating likelihood parameters $\lambdah$ to be specific to each site (indexed by $h$).
\begin{align}
    p( \{ \yh_{1:T}, \fh_{1:T} \}_{h=1}^H )
    &= \prod_{h=1}^H p_{\alpha}( \fh_{1:T} ) p_{\lambdah}( \yh_{1:T} \mid \fh_{1:T} )
\end{align}
For simplicity, for the latent-generating distribution we use the AR model described above with window size $W$, where its latent-generating parameters $\alpha = \{ \beta, \sigma \}$ are shared among all sites.

For the count-generating distribution we use a generalized Poisson, so the site-specific parameter is $\lambdah \in [-1, +1]$.

\paragraph{Priors on latent-generating parameters $\alpha$.} 
We use the same priors as for the single-site GAR model:
\begin{gather}
    \beta_0 \sim \text{Normal}(0, 0.1) \\
    \beta_1 \sim \text{Normal}(1, 0.1) \notag \\
    \beta_2 \sim \text{Normal}(0, 0.1) \notag \\
    \ldots \notag \\
    \beta_W \sim \text{Normal}(0, 0.1) \notag \\
    \sigma \sim \text{HalfNormal}(0.1)
\end{gather}

\paragraph{Priors on count-generating parameters $\lambda$.}
We draw each site's parameter in i.i.d. fashion from the same truncated normal prior as before:
\begin{gather}
\lambdah \sim \text{TruncatedNormal}(0, 0.3, \text{lower}=-1, \text{upper}=1), \quad h \in 1, \ldots H
\end{gather}

\paragraph{Posterior sampling for the multi-site model.}

As with the single-site model, for the multi-site model we can obtain $S$ samples from the posterior using modern MCMC posterior sampling methods:
\begin{align}
    \alpha^s, \left\{\lambdash \right\}_{h=1}^H, 
    \left\{ \fpast^{s,h} \right\}_{h=1}^H
    \sim
    p( \alpha, \{\lambda^h\}_{h=1}^H, \{ \fpast^h \}_{h=1}^H \mid \{\ypast \}_{h=1}^H ), \quad s \in 1, 2, \ldots S.
\end{align}
Then for each site independently, we can draw a \emph{forecast} of future latents 
$\ffuture^{s,h}$ given the past latents for that site and the shared latent-generating parameters $\alpha^s$. If a forecast count sample is needed, we can simply draw $\ysh_{(T+1):(T+F)}$ from the relevant site-specific likelihood with sampled parameter $\lambdash$.

In order to evaluate heldout likelihoods, we can compute the count-normalized log likelihood using the Monte Carlo estimates in Eq.~\eqref{eq:heldout_likelhood_montecarlo}, using $S$ samples of the  site-specific parameters $\lambdah$ and site-specific latents $\ffuture^{h}$.

\end{document}